\documentclass[runningheads]{llncs}
\usepackage{graphicx}
\usepackage[caption=false]{subfig}
\usepackage{siunitx}
\usepackage[pagebackref=false,breaklinks=true,colorlinks=true,bookmarks=false]{hyperref}
\usepackage{amsmath}
\usepackage{amssymb}
\begin{document}
\title{Using Robust Regression to Find Font Usage Trends}
%
%
\author{Kaigen Tsuji\inst{1} \and
Seiichi Uchida\inst{1}\orcidID{0000-0001-8592-7566} \and
Brian Kenji Iwana\inst{1}\orcidID{0000-0002-5146-6818}}
\authorrunning{K. Tsuji et al.}
%
\institute{Kyushu University, Fukuoka, Japan 
\email{kaigen.tsuji@human.ait.kyushu-u.ac.jp
\{uchida, iwana\}@ait.kyushu-u.ac.jp}}
\maketitle              
\begin{abstract}
Fonts have had trends throughout their history, not only in when they were invented but also in their usage and popularity. In this paper, we attempt to specifically find the trends in font usage using robust regression on a large collection of text images. We utilize movie posters as the source of fonts for this task because movie posters can represent time periods by using their release date. In addition, movie posters are documents that are carefully designed and represent a wide range of fonts. To understand the relationship between the fonts of movie posters and time, we use a regression Convolutional Neural Network (CNN) to estimate the release year of a movie using an isolated title text image. Due to the difficulty of the task, we propose to use of a hybrid training regimen that uses a combination of Mean Squared Error (MSE) and Tukey's biweight loss. Furthermore, we perform a thorough analysis on the trends of fonts through time.


\end{abstract}
\section{Introduction}
After Gutenberg invented the printing press with metal types in the fifteenth century, a huge variety of fonts have been created. For example, MyFonts\footnote{\url{https://www.myfonts.com/}} provides more than 130,000 different fonts and categorizes them into six types (Sans-Serif, Slab-Serif,  Serif, Display, Handwriting, Script) for making it easy to search their huge collection. Some of them are very famous and versatile (e.g., Helvetica, Futura, Times New Roman) and some are very minor and rarely used.\par
An important fact about fonts is that they have trends in their history. It is well-known that sans-serif fonts were originally called ``grotesque'' when serif fonts were the majority (around the early twentieth century). However, sans-serif fonts are not grotesque for us anymore. Moreover, many sans-serif font styles (neo-grotesque, geometric, humanist) have been developed after grotesque and had big trends in each era. It is also well-known that psychedelic fonts were often used in the 1970s and recent high-definition displays allow us to use very thin fonts, such as Helvetica Neue Ultralight.\par
In this paper, we attempt to find the trends in font usage using a robust regression technique and a large collection of font images. 
More specifically, we first train a regression function $y=f(x)$, where $x$ is the visual feature of a font image and $y$ is the year that the image was created. Then, if the trained regression function can estimate the year of unseen font images with reasonable accuracy, we can think that the function catches the expected trend --- in other words, there were historical trends in the font usage.\par

The values of this attempt are twofold. First, it will give objective and reliable evidence about the font usage trends, which are very important for typographic designers to understand past history. It should be noted that not only the trends (i.e., main-stream fonts) but also the exceptional usages of fonts are also meaningful. Second, it will help to realize an application to support typography designs. If we can evaluate the ``age'' of the font image by using the regression function, it helps the typographers on their selection of appropriate fonts for their artworks.\par
It is easy to imagine that this regression task is very difficult. For example, Helvetica, which was created in the 1950s, is still used today. In addition, many early typographic designers tried to use futuristic fonts, and current designers often use ``retro'' fonts. In short, font usage is full of exceptions (i.e., outliers) and trends become invisible and weak by those exceptions. In addition, non-experts use the fonts without careful consideration and disturb the trends. Moreover, the ground truth (i.e., the year that a font image was created) is often uncertain or not available, especially for old images.\par
\begin{figure}[t]
    \begin{center}
        \includegraphics[width=0.95\textwidth]{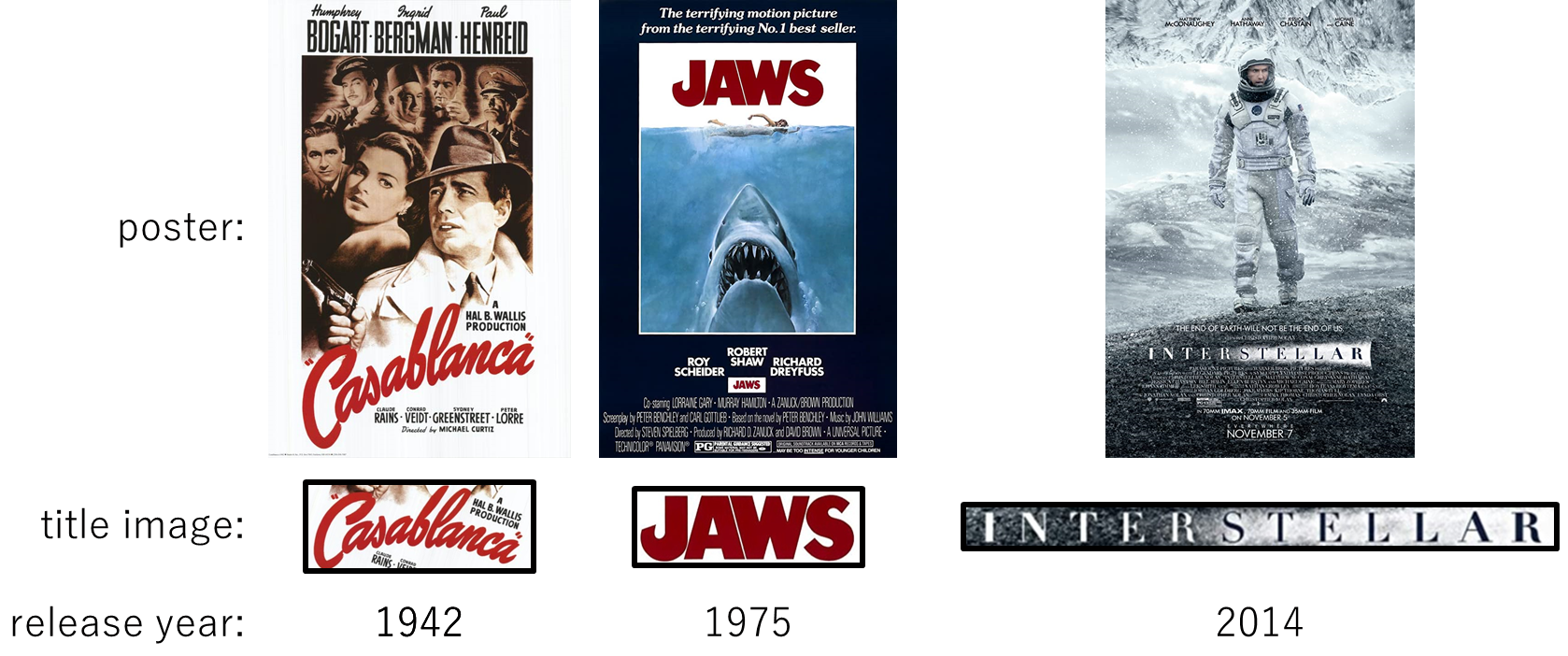}
        \caption{Example movie posters and their title text and release year.} \label{fig:examples_of_movie_posters}
        \vspace{-5mm}
    \end{center}
\end{figure}

 
We tackle this difficult regression task with two ideas. First, we use font images collected from movie posters in this paper, as shown in Fig.~\ref{fig:examples_of_movie_posters}. Since we know when the movie was created and published, we also know the accurate year that the image was made. 
In addition, movie posters are created by professional designers, who are fully careful of the impressions given by fonts and trends. Moreover, the recent movie poster dataset, Internet Movie Database (IMDb), allows us to use 14,504 images at maximum. This large collection is helpful to catch the large trends while weakening the effect of the exceptional usages.\par

Second, we developed a new robust regression method based on deep neural networks with a loss function, Tukey's biweight loss. Deep neural networks realize heavily nonlinear regression functions and thus are appropriate to our difficult task. Tukey's biweight loss has been utilized in robust statistical estimation and, roughly speaking, discounts the large deviations from the average, whereas a typical loss function, such as L2, gives a larger loss for a larger deviation. By this loss function, we can suppress the effect of exceptional samples.\par
One more point that we must consider is what font image features should be used in our year estimation task. Of course, deep neural networks will have a representation learning function, which automatically determines appropriate features. However, it might learn some uninteresting features such as catching the printing quality of old movie posters, like resolution or color fading. Even though the regression might become accurate by these printing quality features, it does not catch the trends of font usages.\par
We, therefore, employ a second feature representation using the font shape features. As for shape features, we use two features; edge images and Scale-Invariant Feature Transform~(SIFT)~\cite{Lowe_1999} features. The former is straightforward; by extracting edges from a font image, it is possible to remove colors and textures that suggest rough printing. The latter is a classical local descriptor and expected to capture the corner and curve shapes of font outlines. By following a standard Bag-of-Visual Words (BoVW) approach, we can have a fixed-dimensional feature representation even though each font image will give a different number of SIFT descriptors.\par
The main contributions of this paper are summarized as follows:
\begin{itemize}
    \item To the authors' best knowledge, this is the first attempt to reveal font usage trends based on regression functions trained on a large number of font images that have the created year as ground truth.
    \item The learned regression function shows a reasonable year estimation accuracy and compares different features as the base of regression.
    \item From a technical aspect, this paper proposes a novel robust regression method based on deep neural networks with Tukey's biweight loss, which helps to realize a regression function robust to exceptional samples (i.e., outliers).  
\end{itemize}

\section{Related Work}
\label{sec:related}

\subsection{Deep regression}

Deep regression is the use of neural networks for regression tasks and it is a wide field of study with many different architectures and applications~\cite{Lathuiliere_2020}. 
Some example applications of deep regression include housing price prediction from house images~\cite{Piao_2019}, television show popularity prediction based on text~\cite{N_2020}, image orientation and rotation estimation~\cite{fischer2015image,Mahendran_2017}, estimation of wave velocity through rocks~\cite{Karimpouli_2019}, stock prices~\cite{Selvin_2017,Lin_2018,Mehtab_2020}, and age prediction~\cite{Niu_2016,Zhang_2019}. 

In these methods, usually, the network is trained using Mean Squared Error~(MSE) or sometimes Mean Absolute Error~(MAE). 
However, using MSE can lead to a vulnerability to outliers~\cite{Belagiannis_2015}. 
Thus, Belagiannis et al.~\cite{Belagiannis_2015} proposed using Tukey's biweight loss function in order to perform deep regression that is robust to outliers. 
Huber loss~\cite{Huber_1964} and Adaptive Huber loss~\cite{cavazza2016active} are other losses that used to be less sensitive to outliers. 
DeepQuantreg~\cite{jia2020deep} is one example of a deep regression model that uses Huber loss. 
However, Lathuiliere et al.~\cite{Lathuiliere_2020} found MSE loss to be better than Huber loss across all four of their tested datasets.

\subsection{Movie poster analysis}

Studying and analyzing design choices of movie posters is of interest to researchers in art and design. 
In particular, research on the relationship between movie poster design and the fonts and text used has been performed. 
For example, Wang~\cite{wang2019art} describes the fonts of movie posters and identified relationships between the tonality, use, and design of the fonts. 
They explain that the fonts on a movie poster contain a large amount of information about the film by communicating artistic expression and symbolic functions to the viewer.
As part of a larger study on movie taglines, Mahlknecht~\cite{Mahlknecht_2015} looked at the relationship between the tagline and poster image. 
In relationship to the design of the tagline text, Mahlkecht found that the size of the tagline on the poster image carries meaning.
Also, many studies have been done that look at the cultural aspects of the design of text in movie posters across the world.
Kim et al.~\cite{kim2019layout} compared the differences between title text (as well as characters, backgrounds, and color contrast) between localized Korean movies and Western movies.
In addition, there have been studies on title design of Indian movies~\cite{Shahid_2014,Shahid_2021} and calligraphy of Chinese movies~\cite{qiang2016calligraphy,Shi_2019}.

The difference between these works and ours is that we take a quantitative approach using machine learning. 
In this regard, there have only been attempts that classify the genre~\cite{chu2017movie,Gozuacik_2019,Sirattanajakarin_2019,kundalia2020multi,Wi_2020,Supriya_2020} and predict box office revenues~\cite{zhou2019predicting}  from movie posters. 
Compared to these, we analyze the movie poster fonts from their relationship with time.



\section{Movie Posters and Title Text}

The elements that make up a movie poster can be roughly divided into two categories: visual information that represents the theme of the movie, and textual information such as the title, actors, production companies, and reviews. The visual information is mainly placed in the center of the poster and conveys the characters in the film and the atmosphere of the film to the viewer. The textual information is often placed at the top or bottom of the poster to avoid overlapping with the visual information. It conveys detailed information to the viewer who is interested in the movie after seeing the visual information. The upper part of Fig.~\ref{fig:examples_of_movie_posters} shows examples of movie posters. Looking at the textual information, the movie title tends to be larger than other textual information (e.g., the names of the director and actors). It is also more elaborately designed than other textual information.\par

In this paper, a title image is defined as a cutout of the part of a movie poster where the title is placed. The title image is a rectangle, and the background information within the rectangle is retained. The lower part of Fig.~\ref{fig:examples_of_movie_posters} shows examples of title images. In some posters, the title is placed vertically or curved which might cause vertical title images or images with significant portions of the background, respectively. 

\begin{figure}[t]
    \begin{center}
        \includegraphics[width=0.7\textwidth]{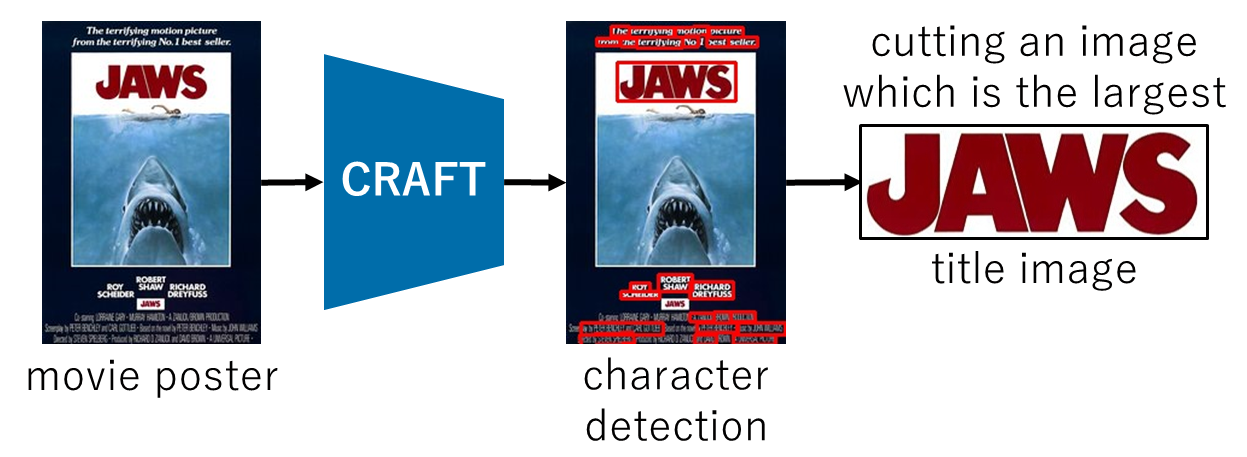}
        \caption{The process of extracting the title image. The text regions on the movie poster are detected using CRAFT~\cite{Baek_2019_Detection} and then the largest area region is cropped to be the title image.} \label{fig:cutting_title_image}
        \vspace{-3mm}
    \end{center}
\end{figure}

To extract the title image, we use a popular scene text detection method called Character Region Awareness for Text Detection~(CRAFT)~\cite{Baek_2019_Detection}. 
Fig~\ref{fig:cutting_title_image} shows the workflow used to create a title image using CRAFT. The poster is input to CRAFT and text regions in the poster are detected then cropped. As mentioned above, the movie titles tend to be larger than other textual information. Thus, the title image is created by cropping the word proposal with the largest area.\par

\section{Year Prediction by Robust Regression}

In this paper, we propose three methods for estimating the year of movie release from title images: Image-based Estimation, Shape-based Estimation, and Feature-based Estimation. In Image-based Estimation, a regression CNN is constructed to estimate the year of movie release given a title image of a movie poster. 
In the case of Shape-based Estimation, edge detection is first applied to the title image in order to remove background and color information. Then a regression CNN is used on the title outline image. 
In Feature-based Estimation, an MLP is constructed to estimate the year of movie release based on SIFT features~\cite{Lowe_1999}. 

\subsection{Image-based Estimation}
\begin{figure}[t]
    \begin{center}
        \includegraphics[width=\textwidth]{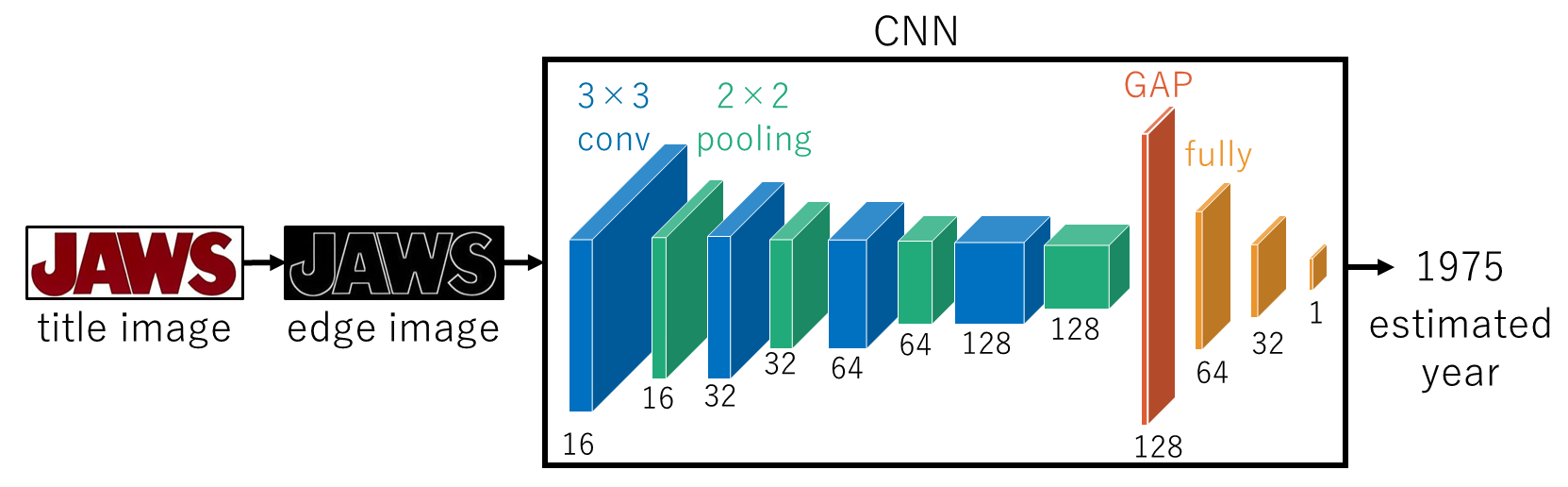}
        \vspace{-3mm}
        \caption{The regression CNN used for Image-based Estimation and Shape-based Estimation.} \label{fig:architecture_of_Image-based_Estimation}
        \vspace{-3mm}
    \end{center}
\end{figure}

In Image-based Estimation, a regression CNN is constructed to estimate the year of movie release when a title image of a movie poster is input. Fig.~\ref{fig:architecture_of_Image-based_Estimation} shows the architecture of the proposed Image-based Estimation. 
The Image-based Estimation regression CNN is a standard CNN except with a single output node representing the year. 

In a standard CNN, a fixed image size is required because of the fixed number of parameters in the fully connected layers. 
In order to input title images of any size to the proposed regression CNN, a Global Average Pooling (GAP) layer~\cite{Lin_2014} is incorporated before the fully connected layer.
The GAP layer outputs the average value of each feature map. Since the number of features output by the GAP layer depends only on the number of feature maps output by the feature extraction layer, title images can be input to the CNN even if they have different sizes.\par

\subsection{Shape-based Estimation}
For Shape-based Estimation, we use edge detection in a pre-processing step before using the same structure as Image-based Estimation (Fig.~\ref{fig:architecture_of_Image-based_Estimation}). 
The reason for using edge images as inputs is to remove the color information from the title image. For the purpose of this paper, it is not desirable that features other than font affect the estimation of the movie release year. 
When raw pixels are used, it is possible for color to have more effect on the regression CNN than the font itself.
By using edge images without color information as input, we attempt to estimate the movie release year based on font features alone. 
It also removes the background image and noise.\par

\subsection{Feature-based Estimation}

\begin{figure}[t]
    \begin{center}
        \includegraphics[width=0.75\textwidth]{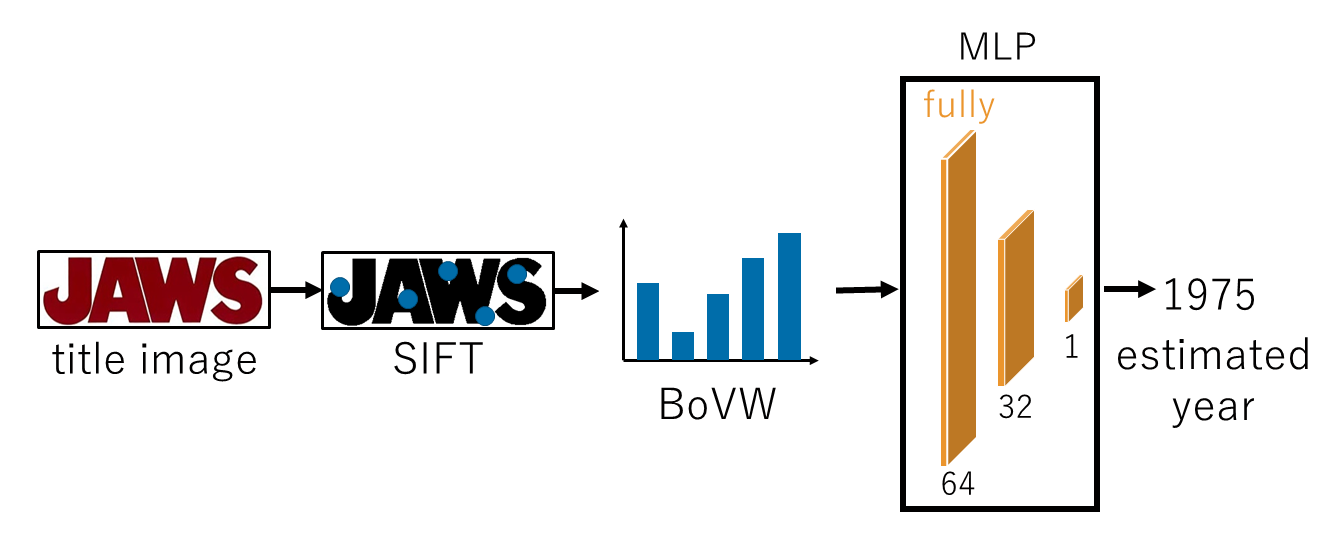}
        \vspace{-3mm}
        \caption{The procedure for performing Feature-based Estimation.} \label{fig:architecture_of_Shape-based_Estimation}
    \end{center}
\end{figure}

In Feature-based Estimation, an MLP is constructed to estimate the movie release year when the local features of the title image are input. Fig.~\ref{fig:architecture_of_Shape-based_Estimation} shows the architecture of Feature-based Estimation. 
SIFT features are extracted from the title image. 
Next, the SIFT features are embedded into a Bag-of-Visual-Words (BoVW) representation, where the bags are found through $k$-means clustering. 

We use SIFT as a method to extract local features from title images. SIFT, which can detect rotation- and scale-invariant feature points, is effective for feature extraction of character designs with various inclinations and sizes. Each detected feature point is represented by a 128-dimensional vector, but since the number of feature points varies depending on the image, it cannot be directly inputted to the MLP.\par

By converting the local features to a BoVW representation, we can make the input data have the same number of dimensions. The method of converting local features to BoVW representation is as follows. First, the local features of all images are divided into $k$ number of clusters using the $k$-means method.  Next, the feature points in each title image are grouped into the nearest cluster and a histogram for each image is created. 
The $k$-dimensional vector created by this process is the BoVW representation.\par

The possibility of removing color information from the title image is another reason for using the BoVW representation as input. As mentioned above, there is a correlation between the font color and the movie release year, but the color information is not included in the features extracted by SIFT. By using the BoVW representation converted from these features, we attempt to estimate the movie release year based on font features alone. 

\begin{figure}[t]
    \begin{center}
        \includegraphics[width=0.45\textwidth]{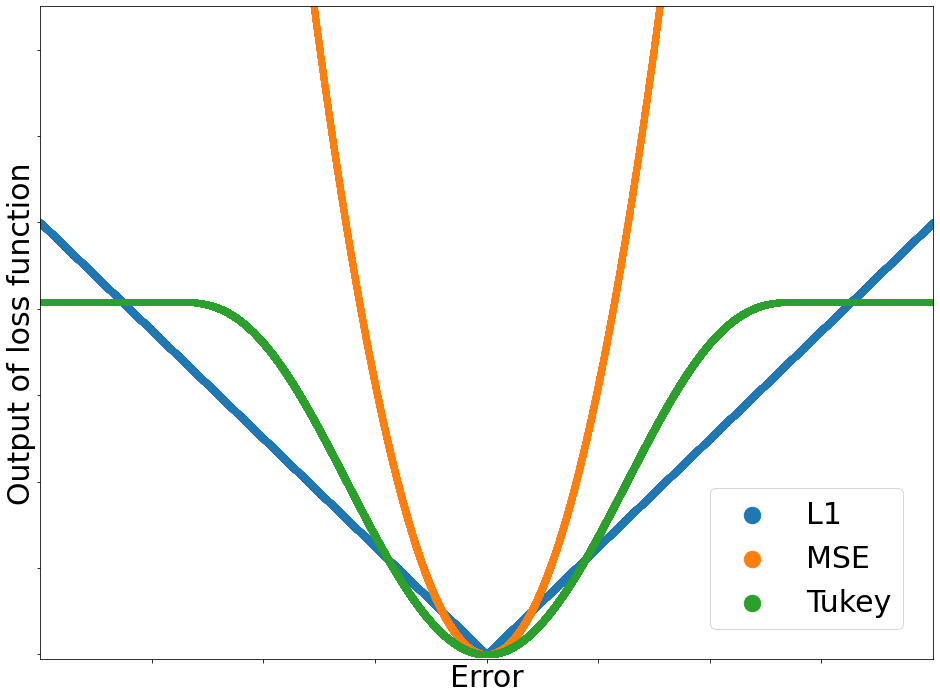}
        \vspace{-3mm}
        \caption{A comparison of different loss functions.} \label{fig:the_plot_of_loss_functions}\vspace{-3mm}
    \end{center}
\end{figure}

\subsection{Robust Regression}

Since there are many variations in designs, some title images have feature values that are far from the distribution in the feature space. For example, some of the designs are bizarre, such as using objects instead of simple lines to compose the text. Such data can be regarded as outliers because it is difficult to estimate the movie release year and the residuals are large.\par

In some cases, the designer intentionally chooses a design that matches the setting year of the movie. For example, the movie release year is in the 2000s, but the poster image uses a retro font design reminiscent of the 1950s to match the setting year of the movie. Such data can also be regarded as an outlier because it is difficult to estimate the movie release year and the residuals are large.\par

To reduce the impact of outliers on learning, we use Tukey's biweight loss, which is a loss function that is robust to outliers~\cite{Belagiannis_2015}. Fig.~\ref{fig:the_plot_of_loss_functions} compares a plot of Tukey's biweight loss with other common loss functions. 
The horizontal axis is the residuals of the Ground Truth and Prediction, and the vertical axis is the output value of the loss function. For MSE loss and L1 loss, the output value always increases as the residuals increase, so outliers have a large impact on learning. On the other hand, for Tukey's biweight loss, the outliers do not affect the learning as much because the output value remains constant when the residual exceeds the threshold value.\par

\subsection{Automatic Loss Switch}
In the early stage of training, the output values of the neural network are unstable. At this time, if Tukey's biweight loss is used as the loss function, the residuals become large and many data are treated as outliers. Therefore, the weights are not updated properly and the learning process does not progress.

We propose the use of MSE loss in the early stages of training and change to Tukey's biweight loss after the output of the neural network has stabilized. In order to determine when the output of the neural network becomes stable, it is possible to check the learning curve of training using only MSE loss in advance. However, there are some disadvantages, such as the need to train once and the need to determine heuristically.

Therefore, we propose a method to determine the stability of the output of a neural network and automatically switch the loss function. Specifically, if the validation loss does not improve for a certain number of epochs, the output is judged to have stabilized. However, there is a certain delay between the time when the output is judged to be stable and the time when the loss function is switched. This prevents switching the loss function when the weights of the neural network fall into a local solution.\par

\section{Experimental Results}
\label{sec:results}

\subsection{Dataset}
The movie poster images and release year were obtained from the Internet Movie Database (IMDb) dataset available on Kaggle\footnote{\url{https://www.kaggle.com/neha1703/movie-genre-from-its-poster}}. For the purpose of language unification, we focused on movies released in the U.S. and obtained data for 14,504 movies. Of these, we selected movies released during the 85-year period from 1932 to 2016.
We randomly selected 56 movies from each year to make the final dataset. The dataset was randomly divided into 20 training images, 8 validation images, and 28 test images. The total dataset consisted of 1,700 training images, 680 validation images, and 2,380 test images.\par

\subsection{Architecture and Settings}
In Image-based Estimation and Shape-based Estimation, Regression CNNs were constructed with four convolutional and pooling layers, one Global Average Pooling (GAP) layer, and three fully-connected layers. The filter size of all convolutional layers was set to $3\times3$, stride 1. The number of filters was 16, 32, 64, and 128 respectively, and Rectified Linear Units (ReLU) were used as the activation functions. Max pooling was used for all pooling layers, and the filter size was $2\times2$ with a stride of 2. The GAP layer was added after the last pooling layer and it computes the average value for each channel, so the number of channels is 128. The number of nodes in the fully-connected layers is 64, 32, and 1, and ReLU is used as the activation function. Dropout was used after the first two fully-connected layers with a dropout rate of 0.5.\par

In Feature-based Estimation, the maximum number of local features to be extracted from the title images using SIFT was 500. The number of clusters $k$ was set to 128 and clustering was performed using the $k$-means method using only the local features extracted from the training images. Then, all title images were converted to BoVW representation. Regression MLPs were constructed with three fully-connected layers. The number of nodes in all the fully-connected layers is 64, 32, and 1, respectively, and ReLU is used as the activation function. Dropout was used after the first two fully-connected layers with a dropout rate of 0.5.\par

The learning procedure is as follows. The number of epochs was set to 3,000. The batch size was set to 128. 
For the proposed loss function, \textit{MSE+Tukey (Prop)}, we used MSE loss in the early stage of learning and Tukey's biweight loss after learning convergence. 
The convergence of learning was judged when the minimum validation loss was not updated for 50 consecutive epochs and automatically switch after 450 epochs if convergence was not reached. We also conducted comparison experiments using only MSE loss, Tukey biweight loss, and a comparative L1 loss. Adam~\cite{kingma2014adam} was used as the optimization method, and the learning coefficient was set to $10^{-4}$. The range of the target variable in the regression, the movie release year was normalized so that the first year starts at 0. 
In other words, we changed the movie release years from 1932-2016 to 0-85, respectively.\par

\begin{table}[t]
    \centering
    \caption{Test results comparison}\vspace{-2mm}
    \begin{tabular}{|c|c||c|c|c|} \hline
        \textbf{Method} & \textbf{Loss function} & \textbf{MAE} $\downarrow$ & \textbf{R$^2$ score} $\uparrow$ & \textbf{Corr.} $\uparrow$ \\ \hline \hline
        Image-based Estimation   & L1 & 16.90 & 0.2472 & 0.5112 \\
        (CNN w/ RGB Images)      & MSE & 17.14 & \textbf{0.2580} & 0.5093 \\
                                 & Tukey & 16.80 & 0.1784 & 0.5079 \\ 
                                 & MSE+Tukey (Prop) & \textbf{16.61} & 0.2002 & \textbf{0.5264} \\ \hline 
        Shape-based Estimation   & L1 & 18.24 & 0.0667 & 0.3793 \\
        (CNN w/ Edge Images)     & MSE & 18.50 & 0.1760 & 0.4209 \\
                                 & Tukey & 18.48 & 0.0389 & 0.3642 \\
                                 & MSE+Tukey (Prop) & 18.35 & 0.0845 & 0.3995 \\ \hline 
        Feature-based Estimation & L1 & 18.78 & 0.0679 & 0.3491 \\
        (MLP w/ SIFT+BoVW)       & MSE & 18.83 & 0.1364 & 0.3712 \\
                                 & Tukey & 18.75 & 0.0434 & 0.3397 \\
                                 & MSE+Tukey (Prop) & 18.71 & 0.0653 & 0.3577 \\ \hline 
        \multicolumn{2}{|c||}{Linear Regression (SIFT+BoVW)} 
      & 19.18 & 0.0768 & 0.3287 \\ \hline
        \multicolumn{2}{|c||}{Constant Prediction} & 21.25 & 0.0000 & 0.0000 \\  \hline
    \end{tabular}
    \label{tab:the_evaluation_indicies}
    \vspace{-3mm}
\end{table}

\begin{figure}
    \centering
    \subfloat[Image, L1 Loss]{
        \includegraphics[clip, width=0.30\columnwidth]{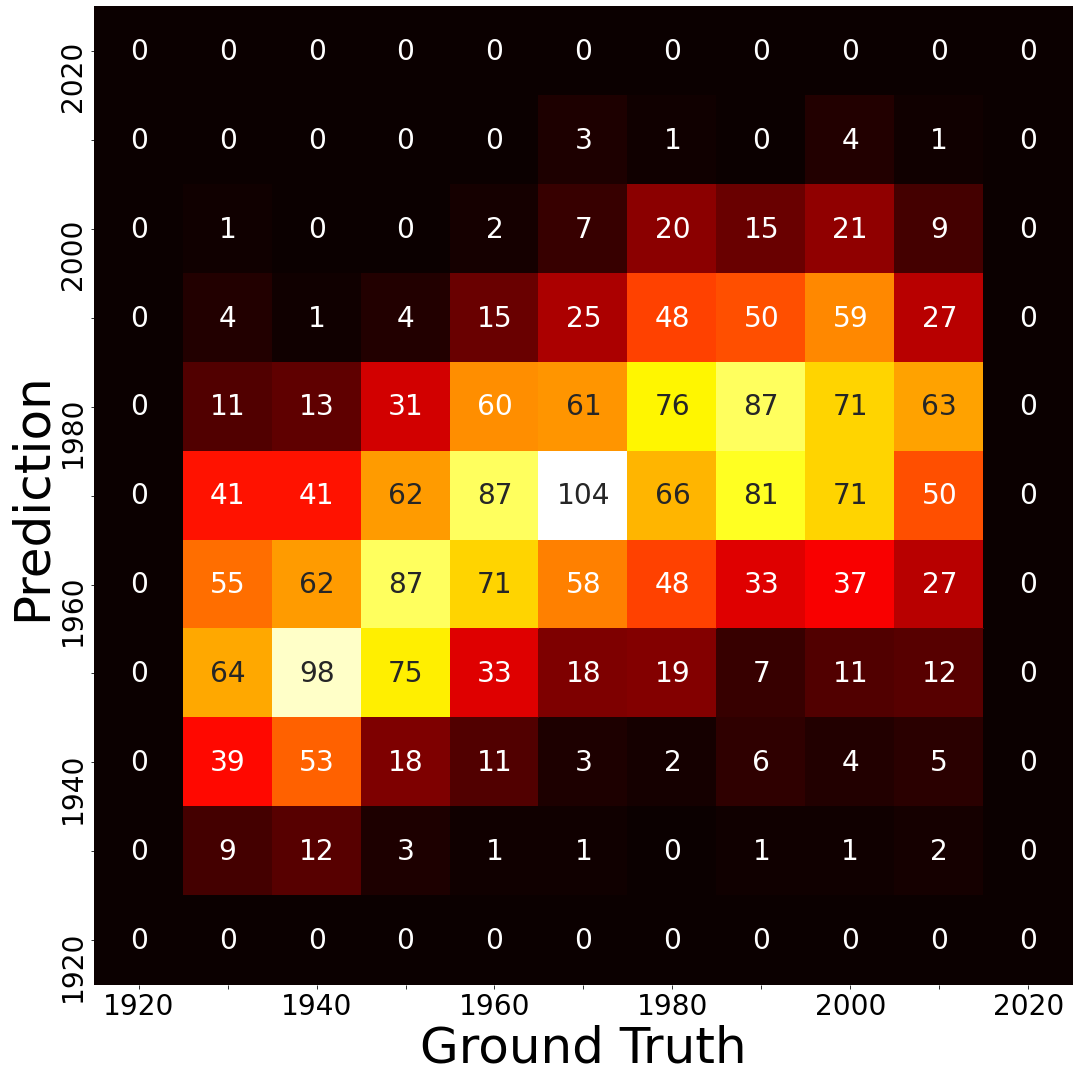}
    }
    \subfloat[Shape, L1 Loss]{
        \includegraphics[clip, width=0.30\columnwidth]{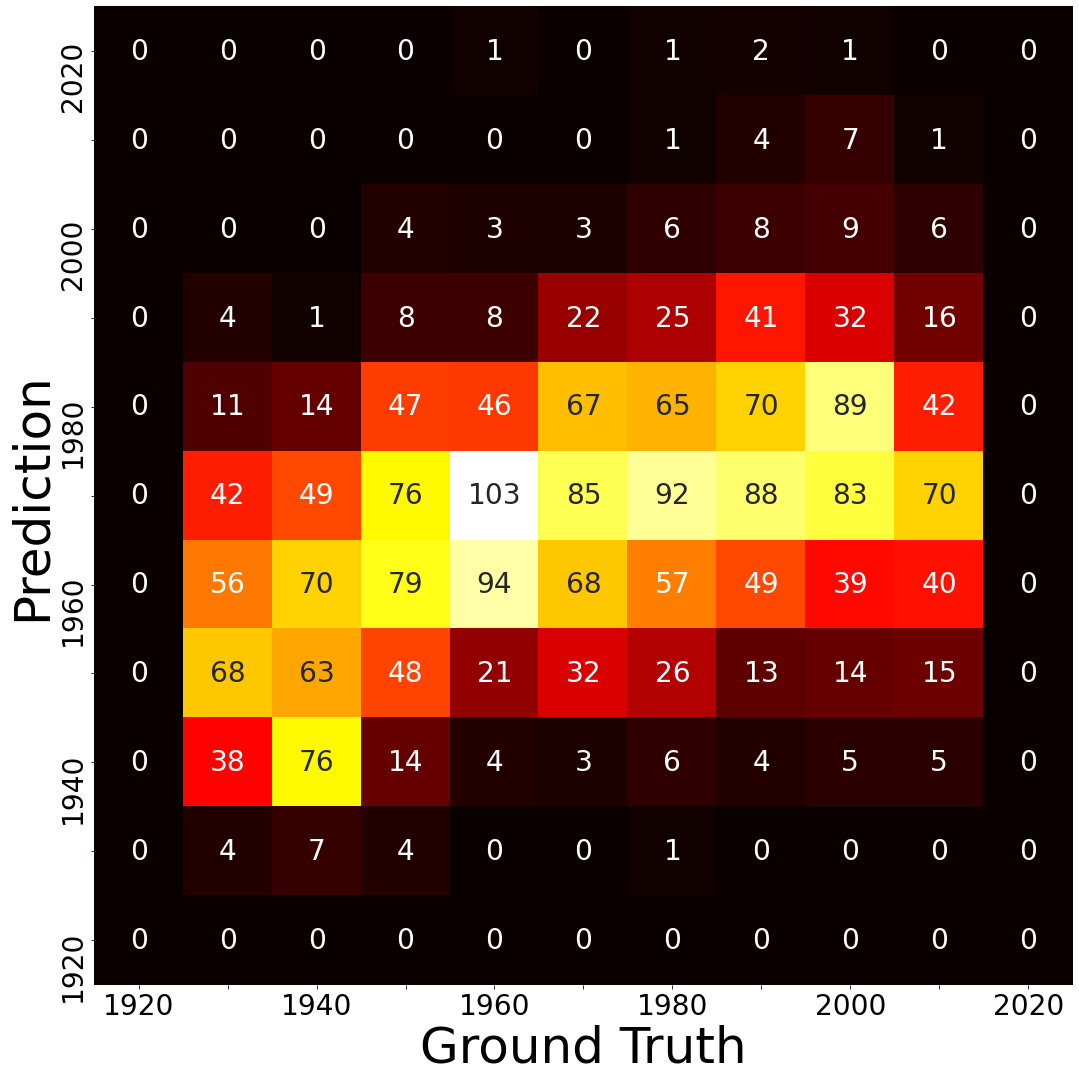}
    }
    \subfloat[Feature, L1 Loss]{
        \includegraphics[clip, width=0.30\columnwidth]{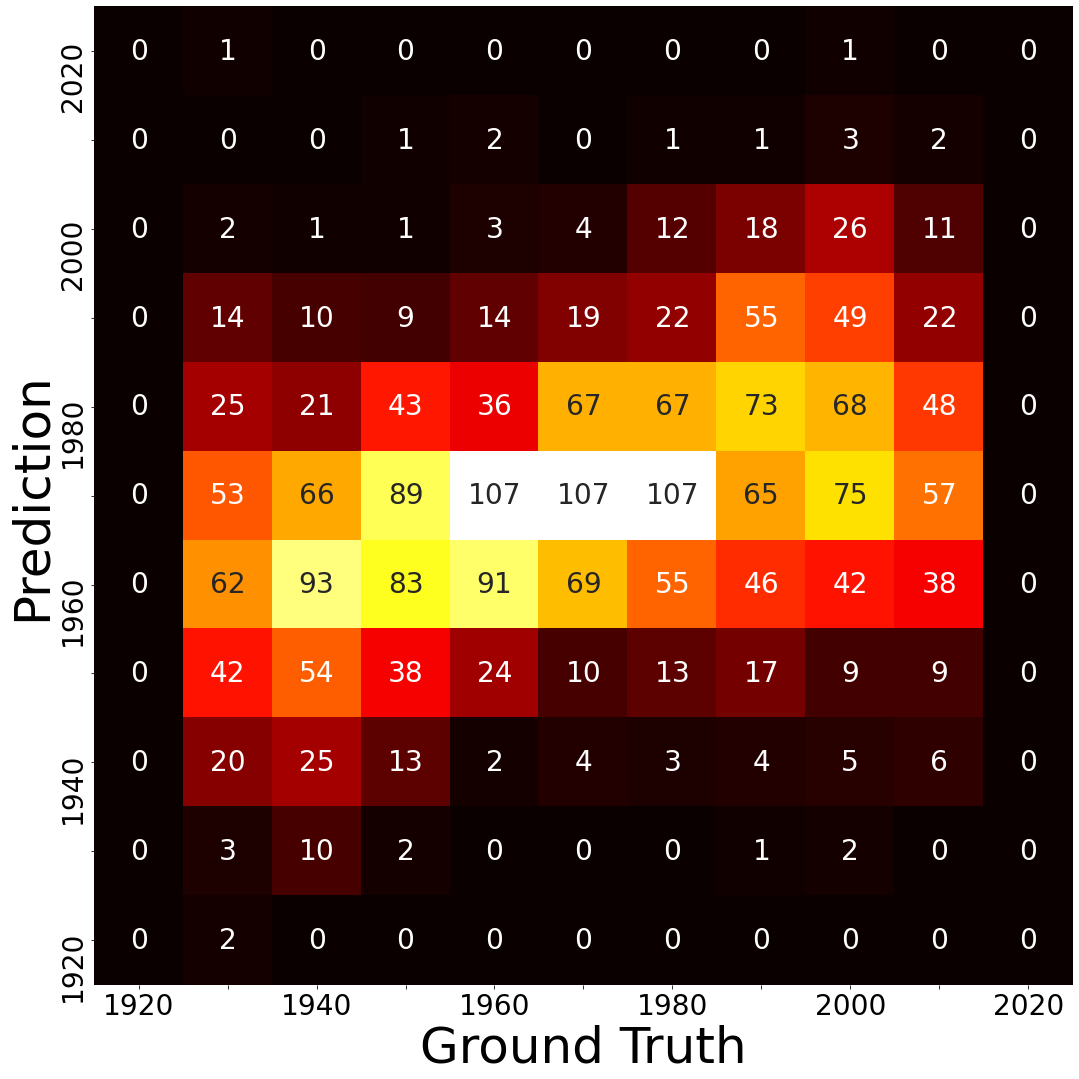}
    }
    
    \subfloat[Image, MSE]{
        \includegraphics[clip, width=0.30\columnwidth]{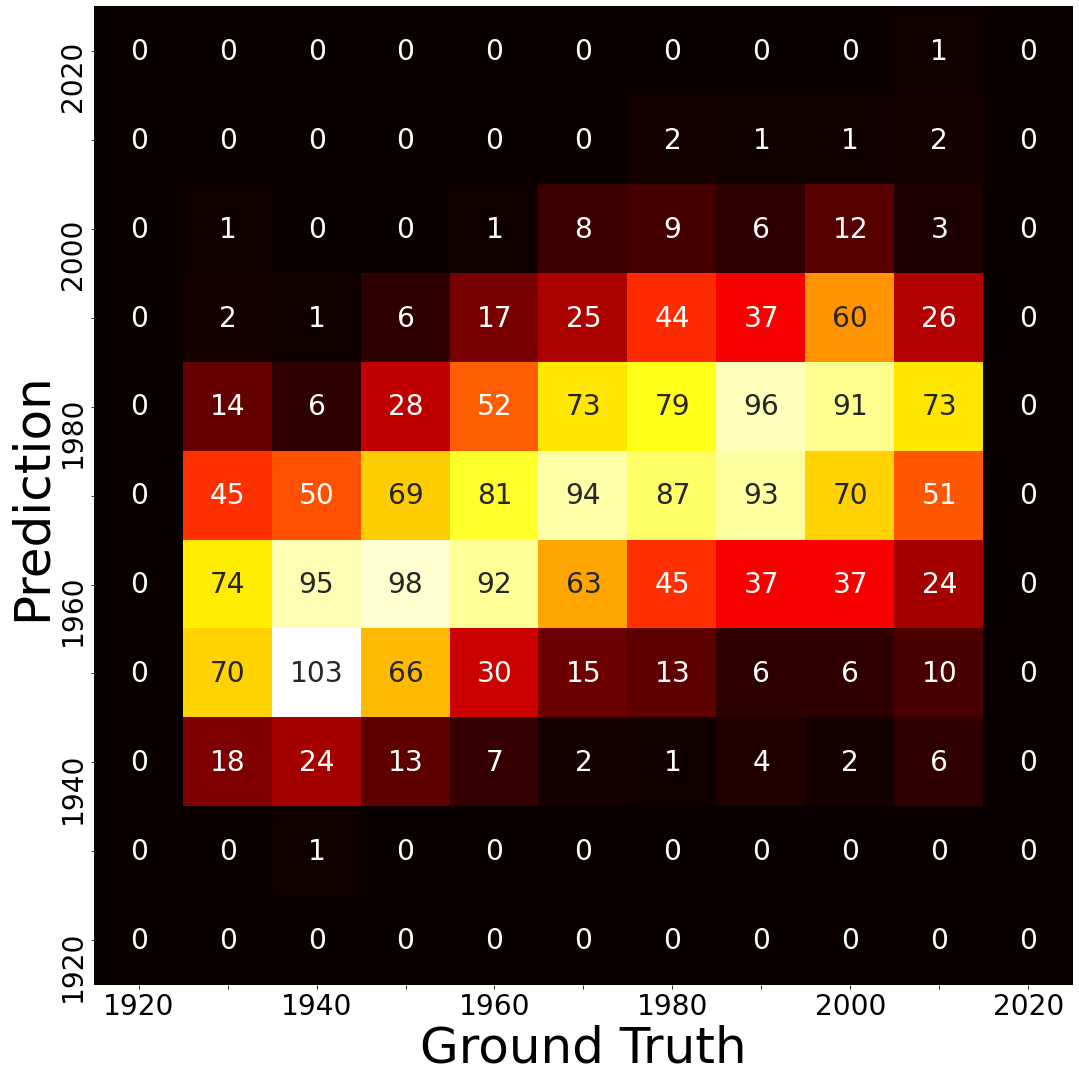}
    }
    \subfloat[Shape, MSE]{
        \includegraphics[clip, width=0.30\columnwidth]{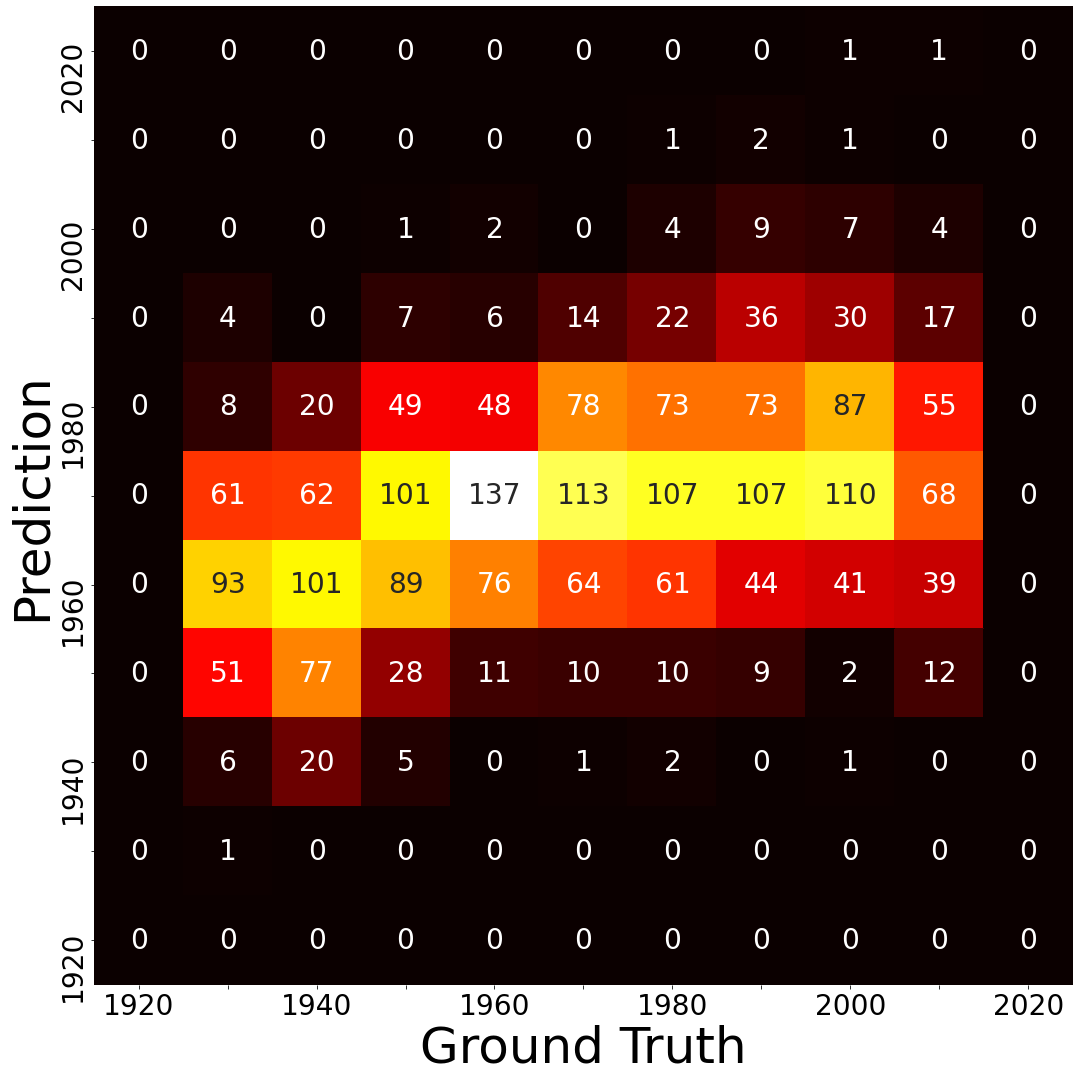}
    }
    \subfloat[Feature, MSE]{
        \includegraphics[clip, width=0.30\columnwidth]{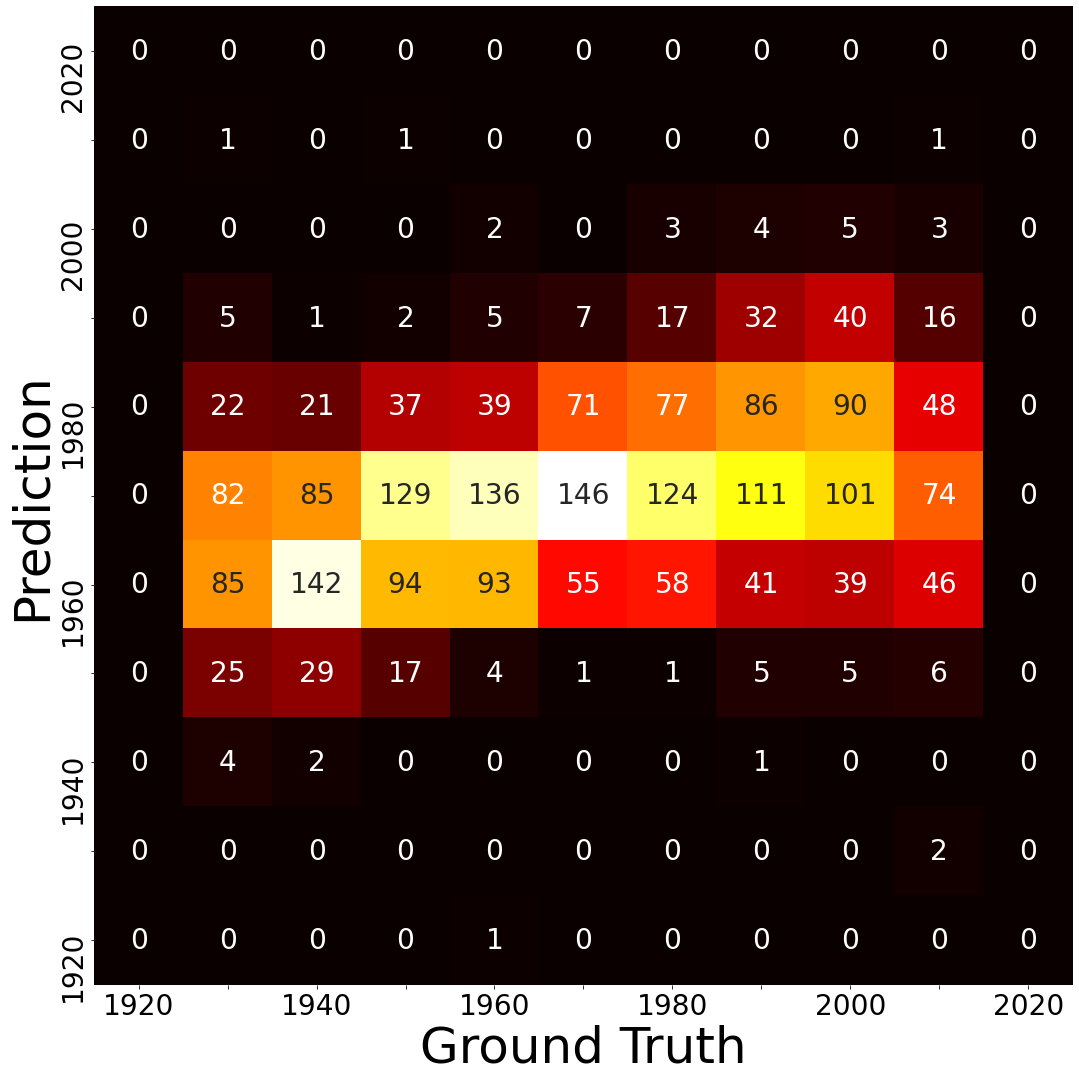}
    }
    
    \subfloat[Image, Tukey]{
        \includegraphics[clip, width=0.30\columnwidth]{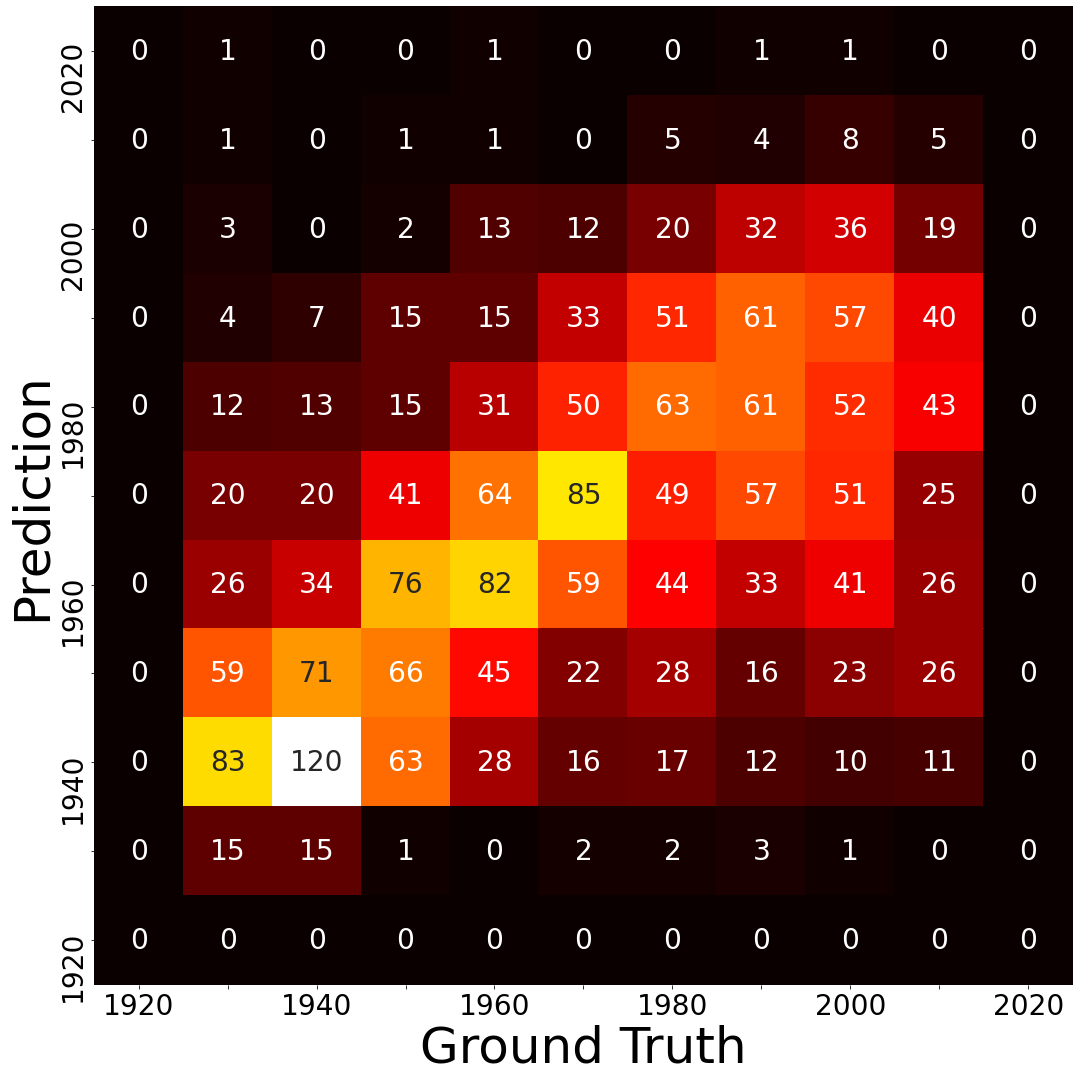}
    }
    \subfloat[Shape, Tukey]{
        \includegraphics[clip, width=0.30\columnwidth]{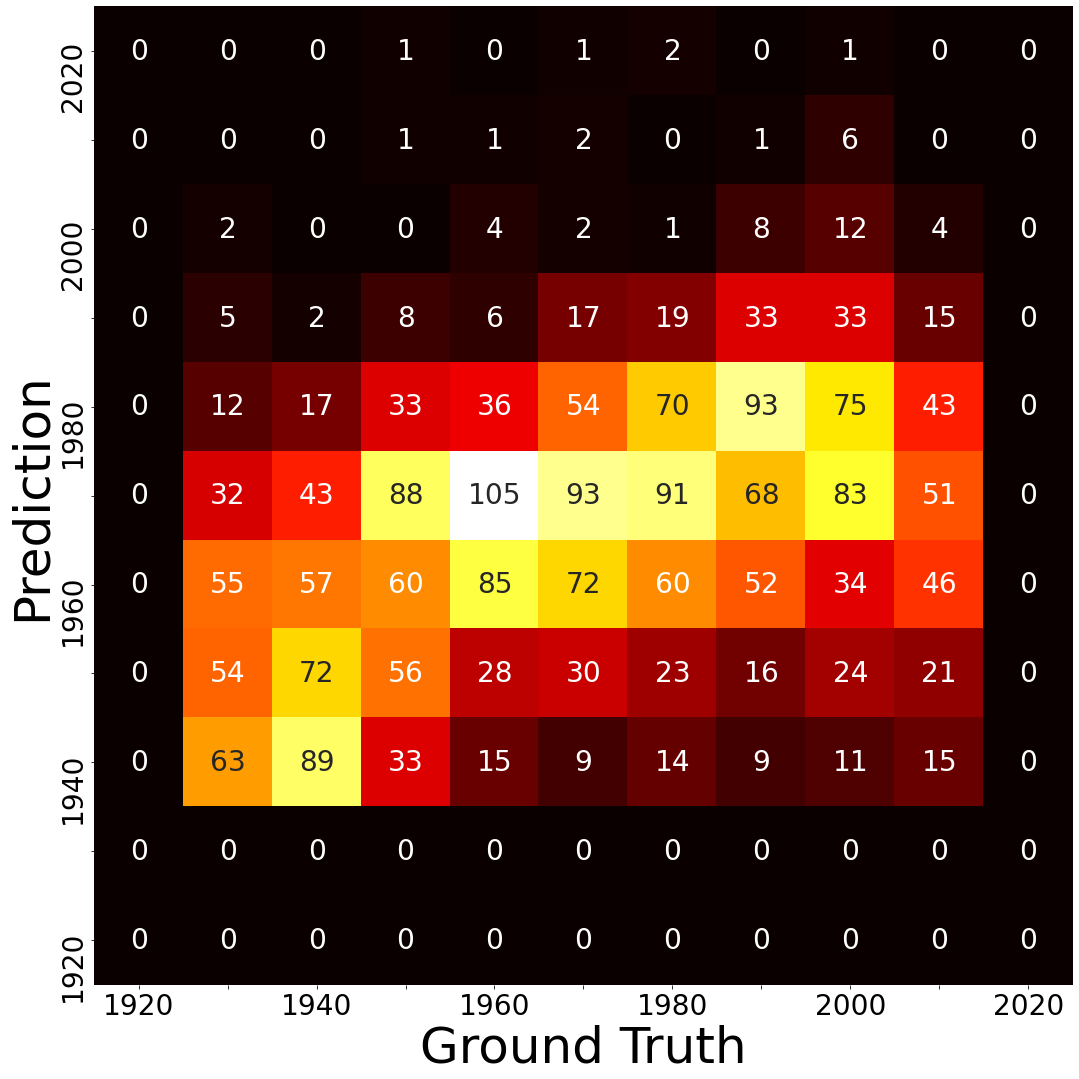}
    }
    \subfloat[Feature, Tukey]{
        \includegraphics[clip, width=0.30\columnwidth]{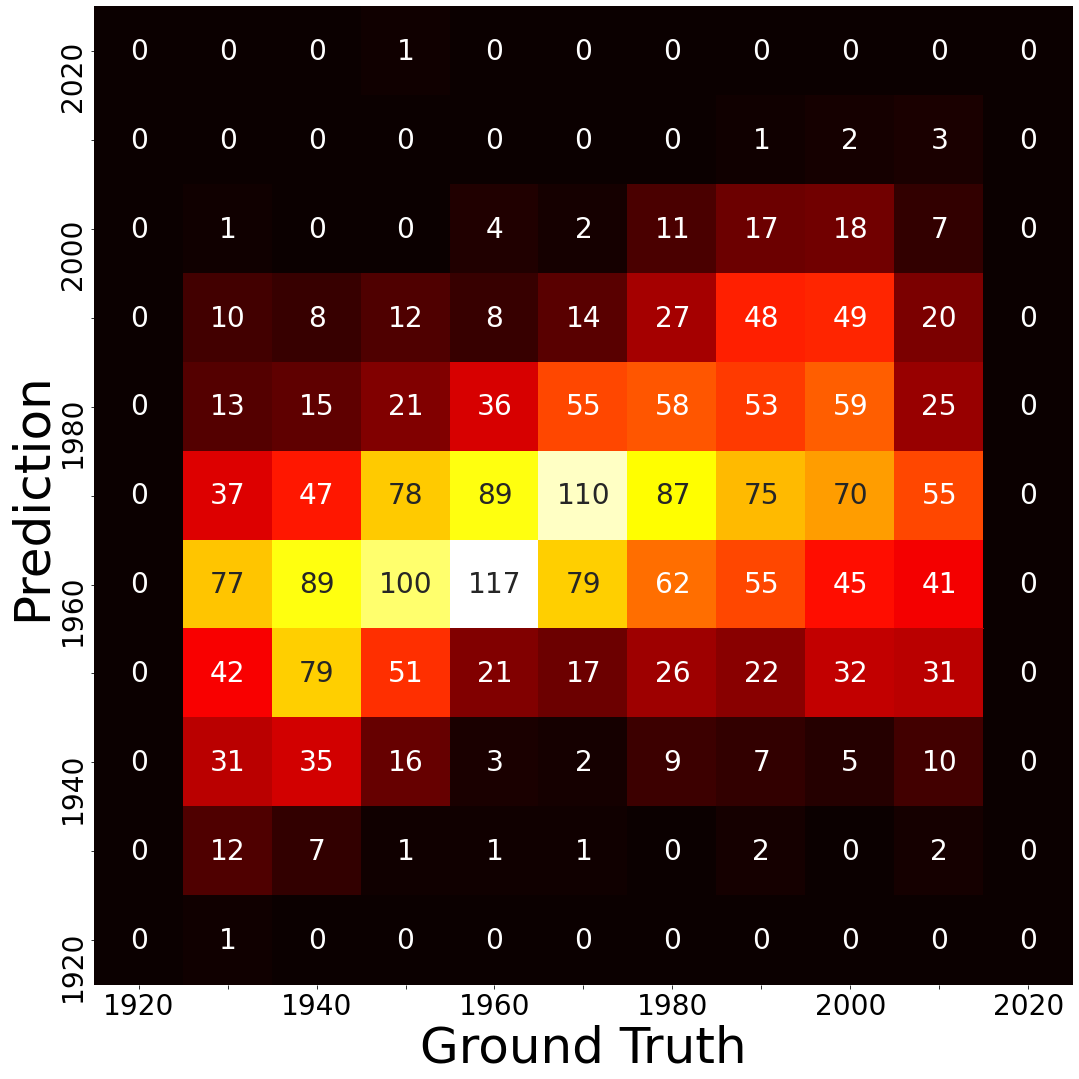}
    }
    
    \subfloat[Image, MSE+Tukey]{
        \includegraphics[clip, width=0.30\columnwidth]{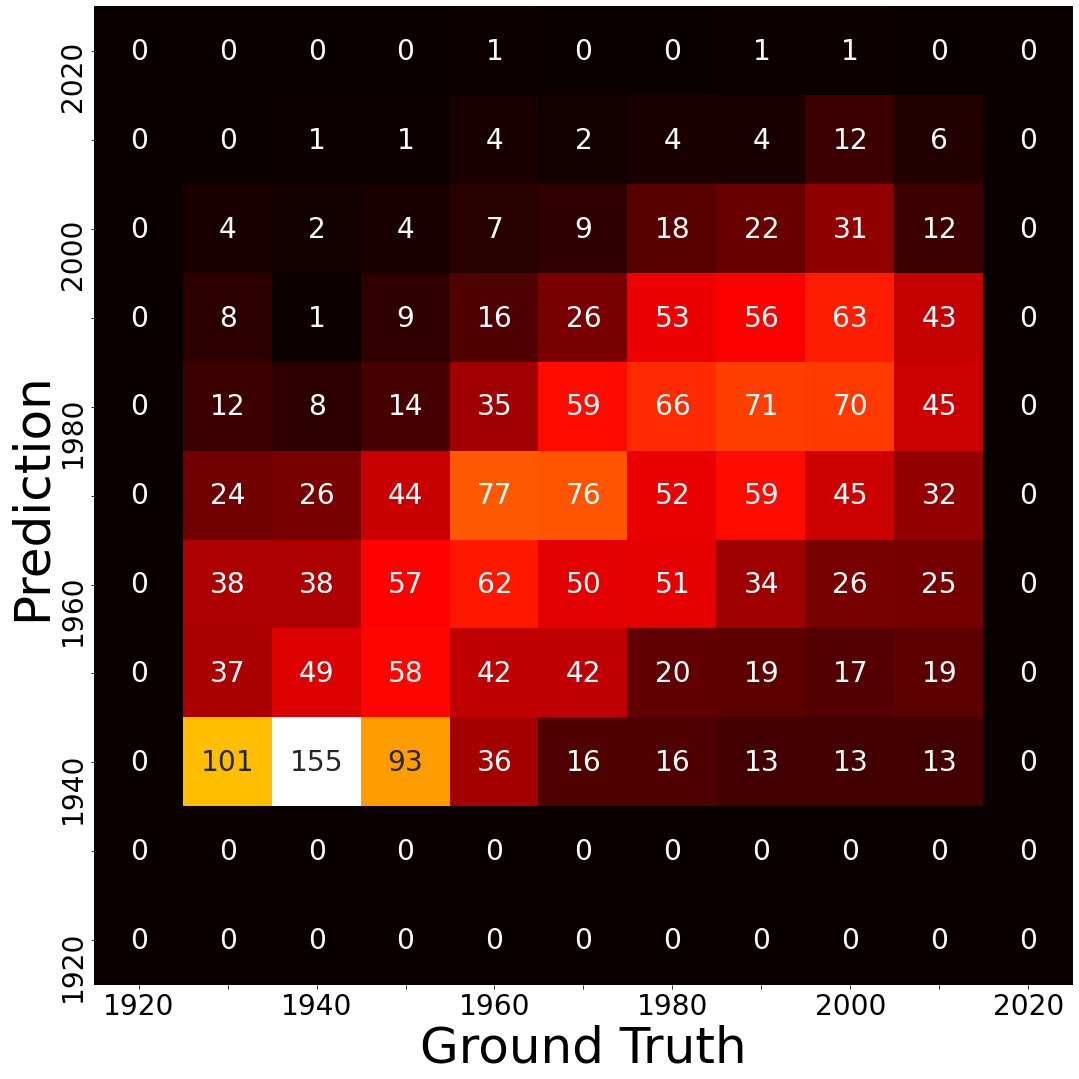}
    }
    \subfloat[Shape, MSE+Tukey]{
        \includegraphics[clip, width=0.30\columnwidth]{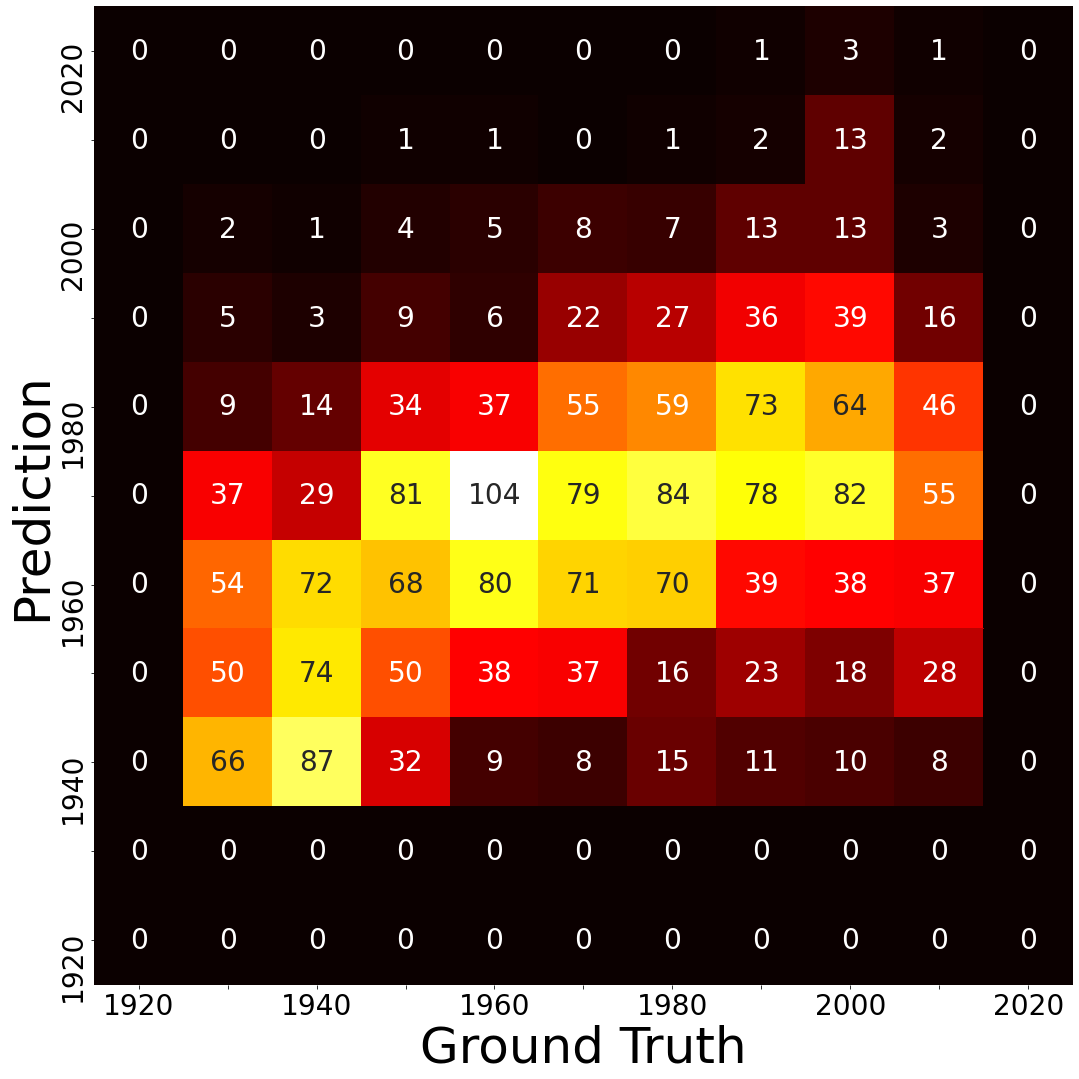}
    }
    \subfloat[Feature, MSE+Tukey]{
        \includegraphics[clip, width=0.30\columnwidth]{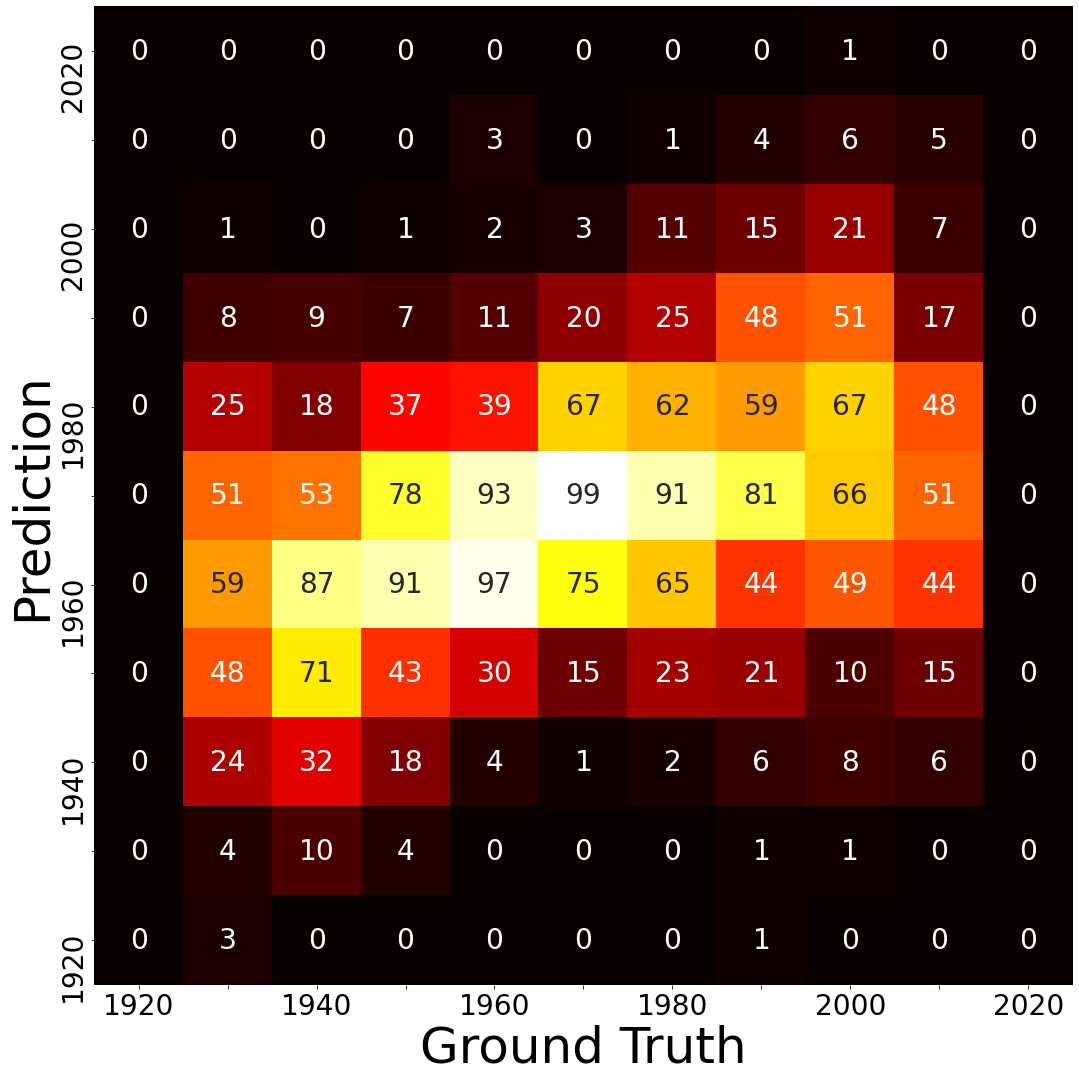}
    }
    \caption{Confusion matrices of each estimation method and loss type.}\label{fig:confusin_matrices}
\end{figure}

\subsection{Results}
To evaluate the proposed method, we use three comparison losses for each estimation type.
The losses used are the \textit{L1} loss, \textit{MSE} loss, \textit{Tukey}'s biweight loss, and the proposed \textit{MSE+Tukey} with the automatic loss switching. 
We also include two more comparisons, \textit{Linear Regression} and \text{Constant Prediction}. 
Linear Regression uses the SIFT features represented by BoVW and Constant Prediction is a baseline where the average year of the training dataset is used for all predictions. 
We evaluated the losses using three metrics, \textit{MAE}, \textit{$R^2$ score}, and the correlation coefficient (\textit{Corr.}).

Table \ref{tab:the_evaluation_indicies} shows the evaluation indices of each method. The performance of Image-based Estimation is better than that of the other methods, indicating that there is a correlation between the color information of the title image and the movie release year. In addition, the performance of Shape-based Estimation is better than that of Feature-based Estimation, indicating that there are other factors in the title image that are related to the year estimation besides local features. 
Linear Regression using the SIFT features in a BoVW representation generally performed worse than the neural network based methods.

When comparing the loss functions, the proposed MSE+Tukey generally performed the best using the MAE metric and MSE loss generally performed the best on the other metrics. 
As mentioned before, this is because MSE loss puts a heavy emphasis on the outliers while Tukey's biweight loss and L1 loss do not. 
Also, in every case, MSE+Tukey performed better than Tukey's biweight loss alone. 
Thus, the benefit of the proposed method is that the loss is a balance between the robustness and emphasis on the outliers. 

Fig. \ref{fig:confusin_matrices} shows confusion matrices of each estimation method and loss type. In all the methods, the prediction is skewed to the center when MSE or L1 loss is used.
When Tukey's biweight loss and the proposed combination of MSE and Tukey's biweight loss is used, the output is spread out much more.
While the evaluation metrics might show that there is an overall worse year estimation, having a wider distribution of predictions is desirable so that analysis can be performed. 
As mentioned before, Tukey+MSE gets the best characteristics of both MSE and Tukey's biweight loss. 

\begin{figure}[t]
    \begin{center}
        \includegraphics[width=0.93\textwidth]{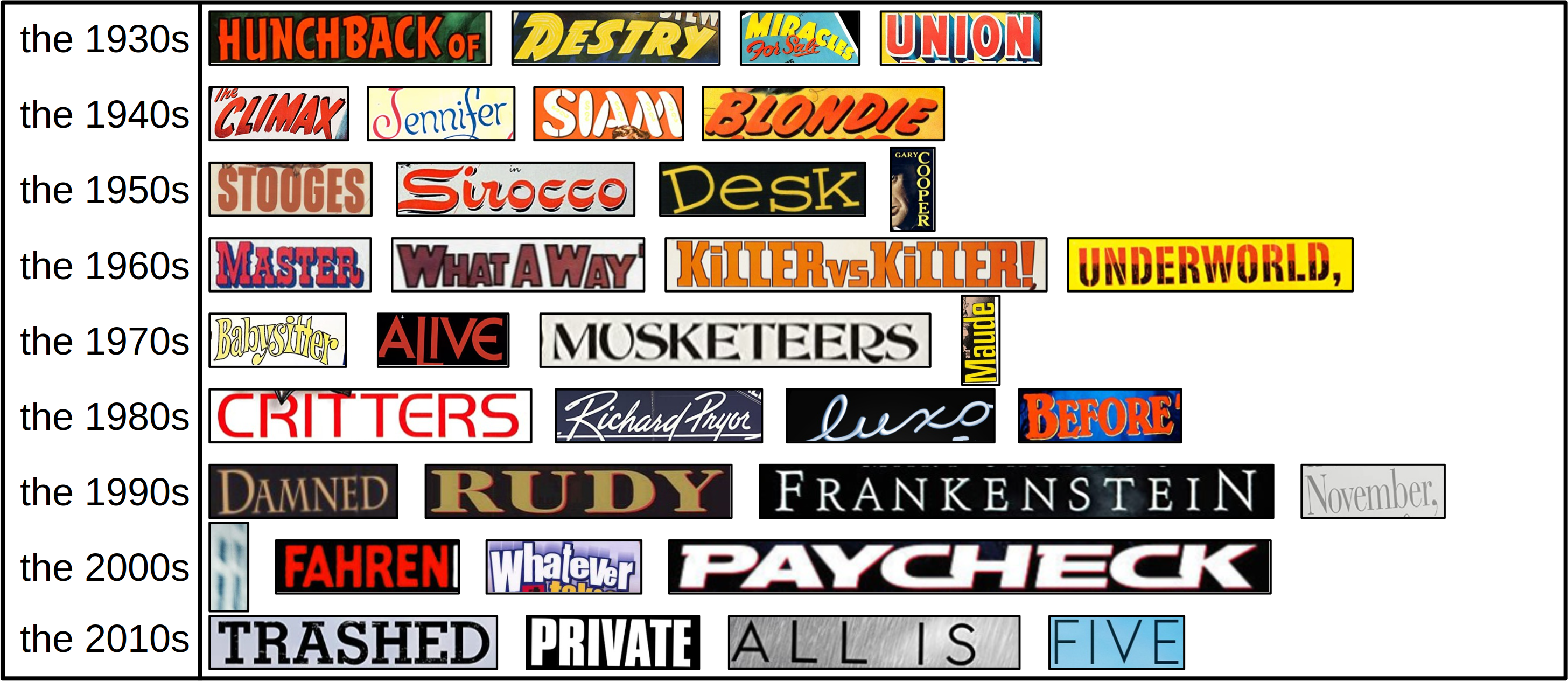}\vspace{-2mm}
        \caption{Image-based Estimation with MSE+Tukey loss: Top four smallest residual title images of each decade. In each case, the year was successfully estimated to the year.} \label{fig:typical_samples_color_msetukey}
        \vspace{-3mm}
    \end{center}
\end{figure}

\subsection{Relationship between color and year}
\begin{figure}[t]
    \centering
    \subfloat[Image-based Estimation]{
        \includegraphics[clip, width=0.93\columnwidth]{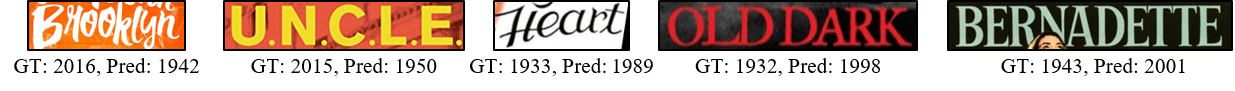}
    }
    
    \subfloat[Shape-based Estimation]{
        \includegraphics[clip, width=0.93\columnwidth]{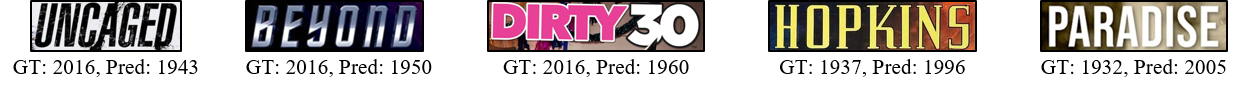}
    }
    
    \subfloat[Feature-based Estimation]{
        \includegraphics[clip, width=0.93\columnwidth]{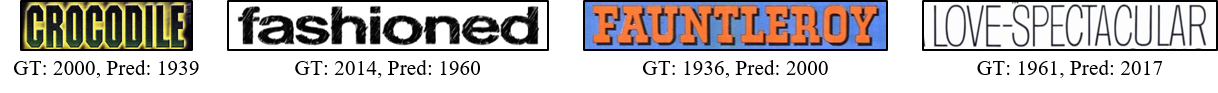}
    }
        \vspace{-2mm}
    \caption{Example title images that have large residuals. Each method uses the proposed MSE+Tukey loss.}\label{fig:exceptional_samples}
    \vspace{-3mm}
\end{figure}

The results from Table~\ref{tab:the_evaluation_indicies} demonstrated that the Image-based Estimation resulted in the highest correlation between the predicted results and the year. 
One reason for this is because the color information in the RGB images conveys a lot of information about the year. 
As shown in Fig. \ref{fig:typical_samples_color_msetukey}, older movie posters tended to have warmer colors, such as red or yellow fonts. 
But, the trend changed around the 1970s to the 1980s, and achromatic and cold colors were used more frequently afterward. 
In addition, dark backgrounds became prominent in the 1980s to the 2000s.
Fig.~\ref{fig:exceptional_samples} shows example title images with large residuals. The ``Brooklyn'' and ``U.N.C.L.E.'' titles, in particular, are warm colors and accordingly were estimated to be older.

\begin{figure}[t]
    \begin{center}
        \includegraphics[width=0.93\textwidth]{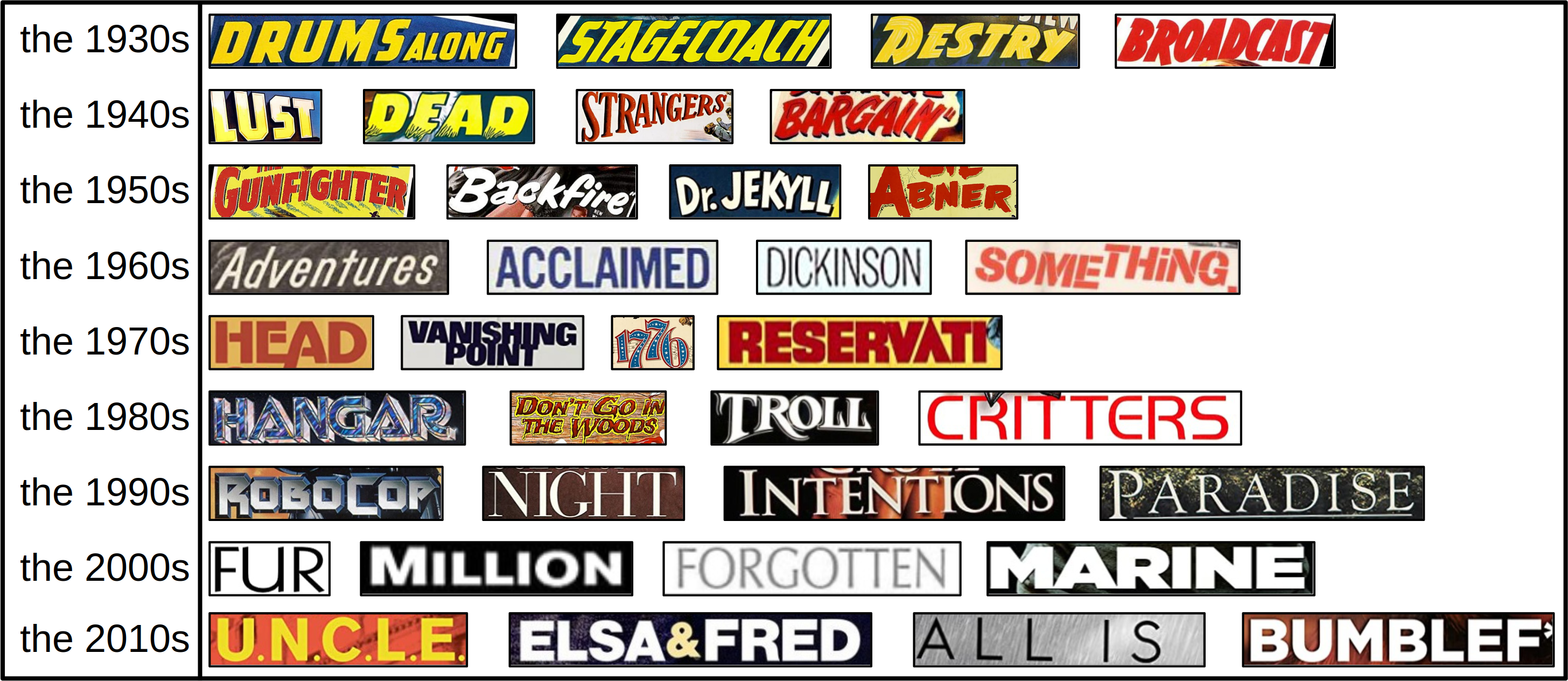}
        \vspace{-2mm}
        \caption{Shape-based Estimation with MSE+Tukey loss: Top four smallest residual title images of each decade. In each case, the year was successfully estimated to the year.}
        \vspace{-3mm} \label{fig:typical_samples_edge_msetukey}
    \end{center}
\end{figure}

\subsection{Relationship between shape and year}
\label{sec:relationship_between_shape_and_year}
\begin{figure}[t]
    \begin{center}
        \includegraphics[width=0.93\textwidth]{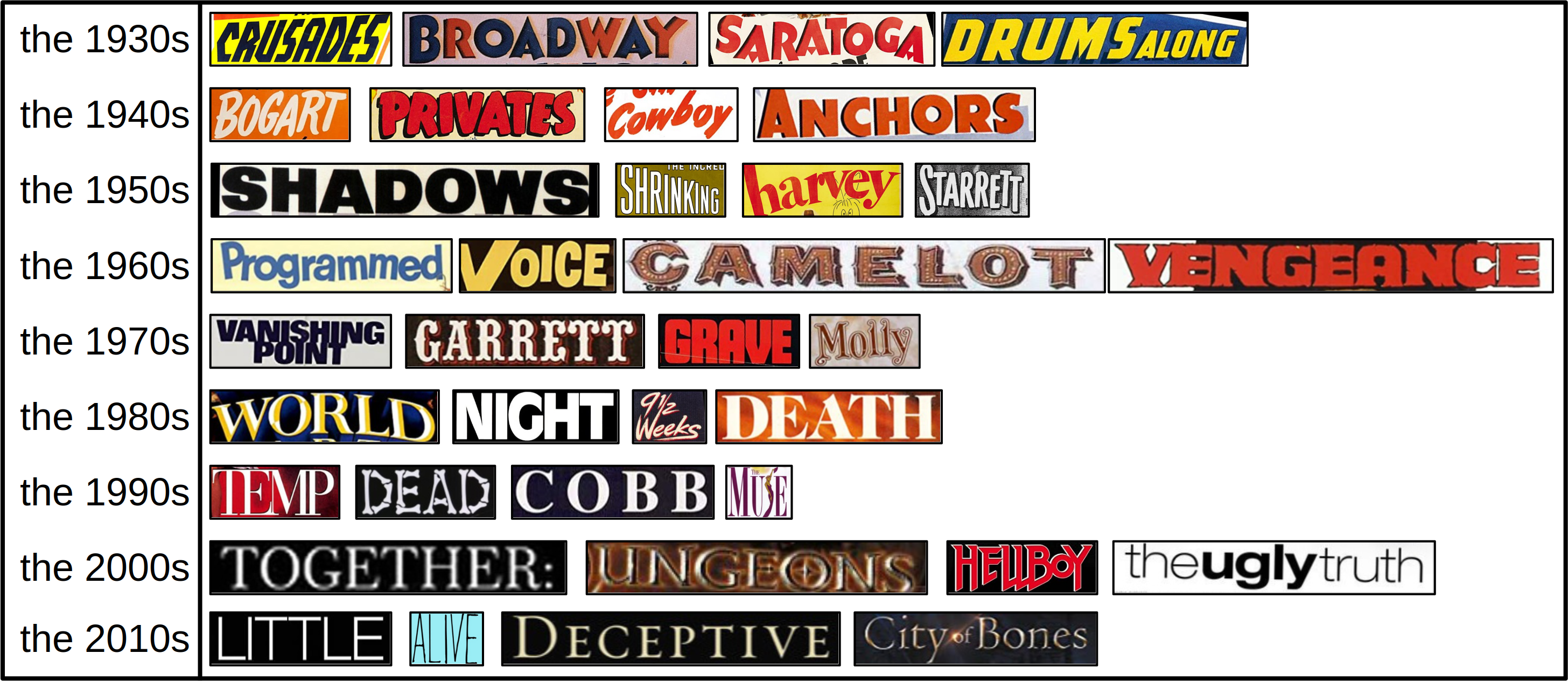}
        \vspace{-2mm}
        \caption{Feature-based Estimation with MSE+Tukey loss: Top four smallest residual title images of each decade. In each case, the year was successfully estimated to the year.}
        \vspace{-3mm}
        \label{fig:typical_samples_bovw_msetukey}
    \end{center}
\end{figure}

Fig. \ref{fig:typical_samples_edge_msetukey} shows the four title images with the smallest residuals for each year in the Shape-based Estimation with MSE+Tukey's biweight loss. In the 1950s and earlier, thicker lines were used more frequently in characters. When examining the presence or absence of serifs, most titles for every decade are sans-serif typefaces, with exception to the 1990s and one from the 1980s. This trend can also be seen in other methods, as shown in Figs. \ref{fig:typical_samples_color_msetukey} and \ref{fig:typical_samples_bovw_msetukey}. This captures the fact that serif fonts
were frequently used for movie posters in the 1990s. 
Another observation that can be found is that for the 1960s and before, the fonts of the movie titles are often skewed, curved, different sizes, or italic. 
This trend is lost after the 1980s, where the fonts are generally straight and uniform.

\subsection{Relationship between local features and year}
Even though the results of Feature-based Estimation performed the worst, there are still clear trends in the SIFT features that can be observed.
As shown in Fig.~\ref{fig:typical_samples_bovw_msetukey}, thicker lines were mostly used before the 1970s, but the trend changed and thinner lines were mostly used after the 1990s. In the 1960s and 1970s, there were also a lot of fonts with decorations. As mentioned in Section \ref{sec:relationship_between_shape_and_year}, in the 1990s, most the title images are serif typefaces. However, unlike Shape-based Estimation, there are still title images with serif typefaces in the 2000s and later when using Feature-based Estimation. 

\section{Conclusion}
\label{sec:conclusion}
In this paper, we attempt to find the trends in font usage using robust regression on a large collection of text images. We use the font images collected from movie posters which are used because they are created by professional designers and portray a wide range of fonts. We found that it is very difficult to find the trends in font usage due to being full of exceptions. We tackle this difficult regression task by a robust regression method based on deep neural networks with a combination of MSE and Tukey’s biweight loss.\par
We experimented using three methods: Image-based Estimation, Shape-based Estimation, and Feature-based Estimation. Using these different estimation methods, we compared L1 loss, MSE loss, Tukey's biweight loss, and the proposed MSE+Tukey with the automatic loss switching. As a result, It becomes clear that the benefit of the proposed method is that the loss is a balance between the robustness and emphasis on the outliers. In addition, It is found that the font usage, such as the thickness, slant of characters, and the presence or absence of serifs, has changed over time. We hope that the findings in this paper will contribute to a better understanding of the historical changes in font design.\par
\bibliographystyle{splncs04}
\bibliography{icdar}
%




\end{document}